\documentclass[pdflatex,sn-mathphys-num]{sn-jnl}

\usepackage{graphicx}%
\usepackage{multirow}%
\usepackage{amsmath,amssymb,amsfonts}%
\usepackage{amsthm}%
\usepackage{mathrsfs}%
\usepackage[title]{appendix}%
\usepackage{xcolor}%
\usepackage{textcomp}%
\usepackage{manyfoot}%
\usepackage{booktabs}%
\usepackage{algorithm}%
\usepackage{algorithmicx}%
\usepackage{algpseudocode}%
\usepackage{listings}%

\theoremstyle{thmstyleone}%
\theoremstyle{thmstyletwo}%

\theoremstyle{thmstylethree}%

\begin{document}

\title[Governance horizon in open-weight AI]{A governance horizon for ethical-use constraints in open-weight AI models}


\author[1,2]{\fnm{Weiwei} \sur{Xu}}\email{xuww@stu.pku.edu.cn}


\author[1,2]{\fnm{Hengzhi} \sur{Ye}}\email{hzye@stu.pku.edu.cn}
\author[3]{\fnm{Haoran} \sur{Ye}}\email{hrye@stu.pku.edu.cn }
\author[4]{\fnm{Kai} \sur{Gao}}\email{kai.gao@ustb.edu.cn}
\author[5]{\fnm{Vladimir} \sur{Filkov}}\email{vfilkov@ucdavis.edu}
\author*[1,2]{\fnm{Minghui} \sur{Zhou}}\email{zhmh@pku.edu.cn}
\affil[1]{\orgdiv{School of Computer Science}, \orgname{Peking University}, \city{Beijing},  \country{China}}

\affil[2]{\orgdiv{Laboratory of High Confidence Software Technologies}, \orgname{Ministry of Education}, \city{Beijing},  \country{China}}

\affil[3]{\orgdiv{School of Intelligence Science and Technology}, \orgname{Peking University}, \city{Beijing},  \country{China}}
\affil[4]{\orgdiv{School of Computer \& Communication Engineering}, \orgname{University of Science and Technology Beijing}, \city{Beijing},  \country{China}}
\affil[5]{\orgdiv{Department of Computer Science}, \orgname{ University of California, Davis},  \city{CA}, \country{ USA}}

\abstract{
Ethical constraints on open-weight AI models are both a reflection of societal concerns and a foundation for AI governance policy. 
The constraints are expected to propagate to downstream derivatives while they are implemented as voluntary metadata disclosures that must be restated at each generation of reuse. We conduct an ecosystem-scale audit of 2,142,823 model repositories on Hugging Face Hub to test whether this disclosure-based governance infrastructure can sustain traceability across deep model lineages. Publicly observable restriction evidence decays with a half-life of 1.31 derivation steps ($R^2 = 0.98$), and beyond seven downstream generations, at least 80\% of descendant models lack sufficient public evidence for a governance determination, a depth boundary we formalize as the governance horizon. Platform-level interventions to restore missing licence metadata reveal that policy design (not enforcement intensity alone) is the binding factor: inheritance-only designs require near-complete enforcement to move the horizon, whereas a mandatory-declaration design that explicitly resolves orphan lineage components shifts the horizon already at moderate enforcement. The structural bottleneck is the existence of lineages with no inheritable upstream intent: such orphan components remain undecidable under any inheritance-only policy regardless of enforcement rate, and unresolved upstream nodes additionally create direct downstream undecidability bottlenecks that inheritance rules alone cannot recover. A comparison with PyPI software dependencies, where governance signals are carried by explicit machine-readable declarations, corroborates that the collapse is topology-specific to open-weight derivation rather than inherent to open ecosystems. These results establish that disclosure-based governance has a shallow, structurally determined reach in open-weight AI, and that achieving deep supply-chain accountability requires provenance mechanisms propagating governance signals through derivation itself.}




\maketitle
\section{Introduction}\label{sec1}
Open-weight AI models, a subset of foundation models with publicly available weights~\cite{bommasani2021opportunities,touvron2023llama}, are increasingly reused, adapted, and recombined through operations such as fine-tuning, adapters, and model merging~\cite{hu2022lora,stalnaker2025empirical,goddard2024arcee,yadav2023ties}.
The very openness that empowers open-weight AI models also exposes them to unprecedented risks of technological abuse.
To mitigate dual-use risks and constrain downstream misuse, upstream model creators often attach ethical-use, responsible-use, or other restrictive terms intended to shape later deployment and reuse~\cite{mcduff2024standardization,contractor2022openrail,meta2024llama3}. However, unlike model capabilities that are often preserved through weight-level derivation, such constraints are not carried forward by derivation itself. Instead, they are expressed primarily through model cards, license fields, and other forms of voluntary disclosure attached to public artifacts~\cite{stalnaker2025empirical,mitchell2019model}. Open-weight ecosystems thus exhibit a fundamental governance mismatch: restrictions are expected to behave as inheritable constraints, but are implemented mainly as optional disclosures, persisting only insofar as successive downstream releases continue to restate them in public-facing artifacts. This mismatch echoes broader concerns about ``data cascades'' in high-stakes AI, where upstream documentation work is structurally undervalued relative to model-level work~\cite{sambasivan2021everyone}.

This mismatch matters because downstream governance depends not only on whether restrictions were ever articulated, but also on whether they remain auditable as models propagate across generations. When restrictive terms become difficult to recover, interpret, or attribute after only a few hops of derivation, compliance and accountability shift from questions of rule interpretation to questions of evidence availability~\cite{longpre2024large}. In such case, a central failure mode of governance is not directly observable violation, but growing undecidability: downstream developers, deployers, and regulators can no longer determine from public artifacts alone which restrictions continue to apply.

This problem is becoming increasingly consequential as open-weight model ecosystems expand and regulatory attention turns to supply-chain traceability, provenance and accountability~\cite{bommasani2023ecosystem}, and as the broader societal implications of generative AI are debated across science and policy venues~\cite{samuelson2023generative,epstein2023art}. Prior work has examined model documentation~\cite{mitchell2019model}, foundation model transparency~\cite{bommasani2023foundation,bommasani2024foundation}, dataset documentation and provenance~\cite{gebru2021datasheets,longpre2024data}, emerging licensing risks in AI supply chains~\cite{wang2026hidden}, and the structure of pre-trained model supply chains on sharing platforms~\cite{jiang2024peatmoss}. However, little is known about how ethical-use constraints propagate through deep open-weight model lineages, whether the public evidence of such constraints systematically degrades with lineage depth, or whether deep auditability collapse is an inevitable by-product of openness, rather than a topology-specific consequence of open-weight derivation. In particular, current governance practice implicitly assumes that disclosure can scale with reuse, yet this assumption has rarely been tested empirically in large lineage graphs. 
Unlike software dependencies, which are typically represented through explicit reference relations, open-weight reuse often proceeds through weight-level transformation, making governance signals dependent on repeated disclosure rather than on derivation alone~\cite{oderinwale2025anatomy}. 

\begin{figure}[h]
\centering
\includegraphics[width=0.99\textwidth]{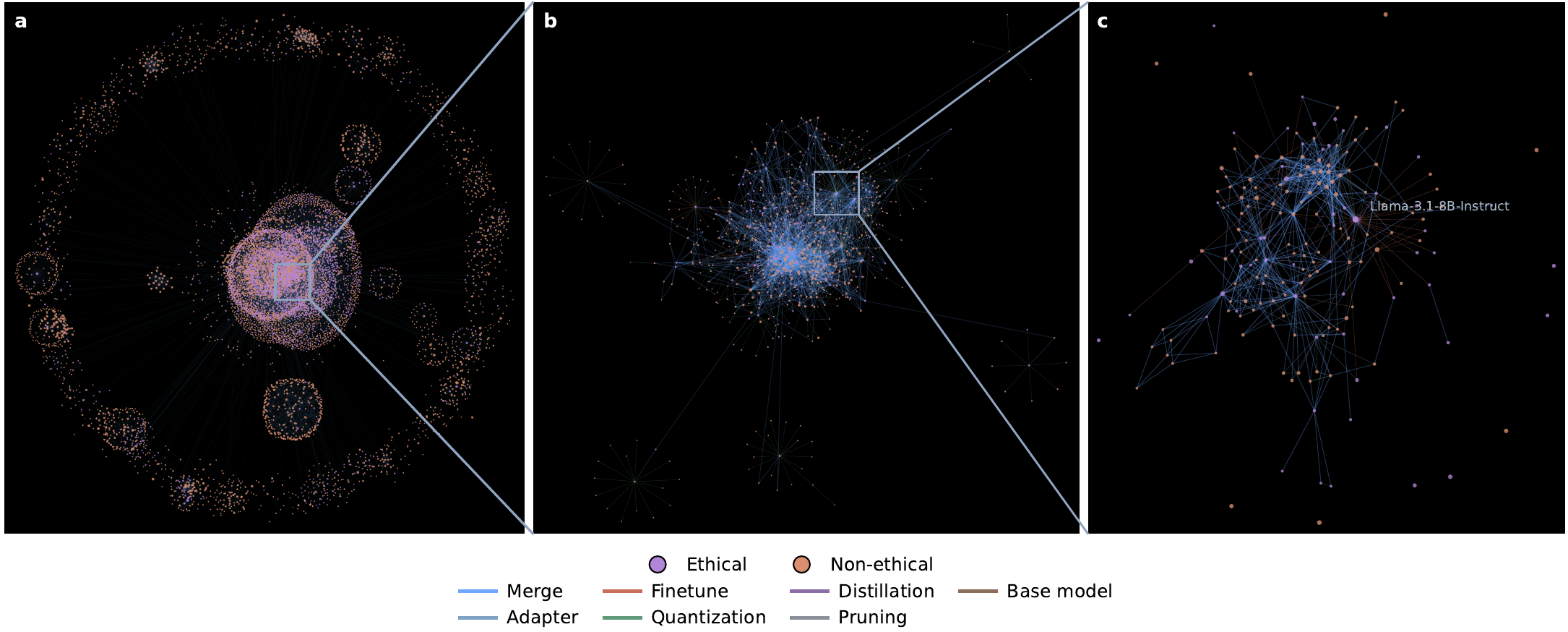}
\caption{Structural topology of open-weight model lineages on Hugging Face. Nodes represent models, and edges represent validated direct derivation relationships extracted from public artifacts. Rather than forming simple tree-like inheritance chains, open-weight reuse produces dense lineage topologies with clustered model families and entangled local neighborhoods around widely reused models, creating the structural conditions through which governance signals must propagate. \textbf{a}, Global view of the validated lineage graph. \textbf{b}, Zoomed view of a local lineage community illustrating clustered reuse and heterogeneous derivation pathways. \textbf{c}, Trimmed neighborhood centered on a widely reused focal model, highlighting how local ancestry remains structurally complex even after graph simplification. In \textbf{b} and \textbf{c}, edge colours indicate derivation operators, including merge, adapter, fine-tuning, quantization, distillation and pruning; edges without a more specific assignment are shown as base-model relationships. Node colours provide an illustrative overlay of whether a model carries publicly observable ethical-use restrictions according to our licence analysis.}\label{fig1}
\end{figure}

Figure~\ref{fig1} illustrates the structural setting in which this governance problem arises. Rather than forming simple tree-like inheritance chains, open-weight reuse on Hugging Face produces dense lineage topologies with clustered model families, repeated downstream transformations, and entangled local neighborhoods around heavily reused models. These topological features make governance signals dependent on repeated public restatement across generations, rather than on technical derivation alone.

Here we conduct a large-scale ecosystem audit of ethical-use constraints across Hugging Face model lineages. We trace how publicly observable restriction evidence changes with lineage depth (hops), distinguish broad auditability loss driven by missing or ambiguous evidence from residual weakening within observable cases, and examine whether merge-induced upstream conflicts are associated with directional shifts toward more permissive declarations. We further compare these patterns with a PyPI software-dependency comparator to isolate properties specific to open-weight evolutionary topology.

We find that publicly observable evidence of ethical-use constraints does not reliably inherit across generations. Instead, restriction evidence declines sharply with lineage depth, while missing and ambiguous cases rapidly dominate downstream audit outcomes. Missing evidence is a major component of this collapse, but not its whole explanation: even where evidence remains present, restriction retention continues to decline, and observable cases show directional shifts toward more permissive declarations. Together, these results allow us to formalize an operational governance horizon: a depth-indexed boundary beyond which metadata-based auditing becomes predominantly statistically undecidable. By showing that this horizon is not a generic artifact of open ecosystems but a topology-specific limit of open-weight derivation, our work identifies a structural depth boundary for disclosure-based governance in open-weight AI and motivates a shift toward more robust, lineage-aware governance mechanisms.

\section{Disclosure-based governance does not scale with open-weight reuse}
Our analysis is based on an ecosystem snapshot of 2{,}142{,}823 public model repositories on Hugging Face, collected in October 2025, from which we extract 1{,}033{,}781 validated model-to-model derivation relationships spanning fine-tuning, adapter, merge, quantization, distillation, pruning and base-model operations (see Section~\ref{Methods} for extraction and validation details). For each repository, we classify the associated licence evidence as carrying ethical-use restriction signals or not, using a rule-based heuristic classifier evaluated against a human-adjudicated gold-standard set (precision 0.96, recall 0.91; Section~\ref{Methods}). We define source models as lineage roots (in-degree zero) that carry ethical-use restriction evidence, yielding 1{,}916 sources and 133{,}812 reachable descendants within 10 downstream hops. To characterise how governance signals change with lineage depth, we classify each downstream model into one of four audit states, Decidable, Inconsistent, Undecidable--Missing (UM) and Undecidable--Ambiguous (UA), based on the publicly available licence evidence at the model and its upstream ancestors (Section~\ref{Methods}). Dataset dependencies are excluded throughout, as the present study concerns model-to-model propagation of governance signals.

\subsection{Restriction evidence does not reliably inherit across generations}\label{sec:restriction-retention}

We begin by tracing the downstream retention of ethical-use restriction evidence from ethical source models. Retention at hop~$h$ is measured as the proportion of downstream descendants that continue to carry publicly observable ethical-use restriction evidence. This binary measure captures whether ethical-use constraints remain visible as models are repeatedly adapted and redistributed across successive generations.

\begin{figure}[t]
\centering
\includegraphics[width=0.99\textwidth]{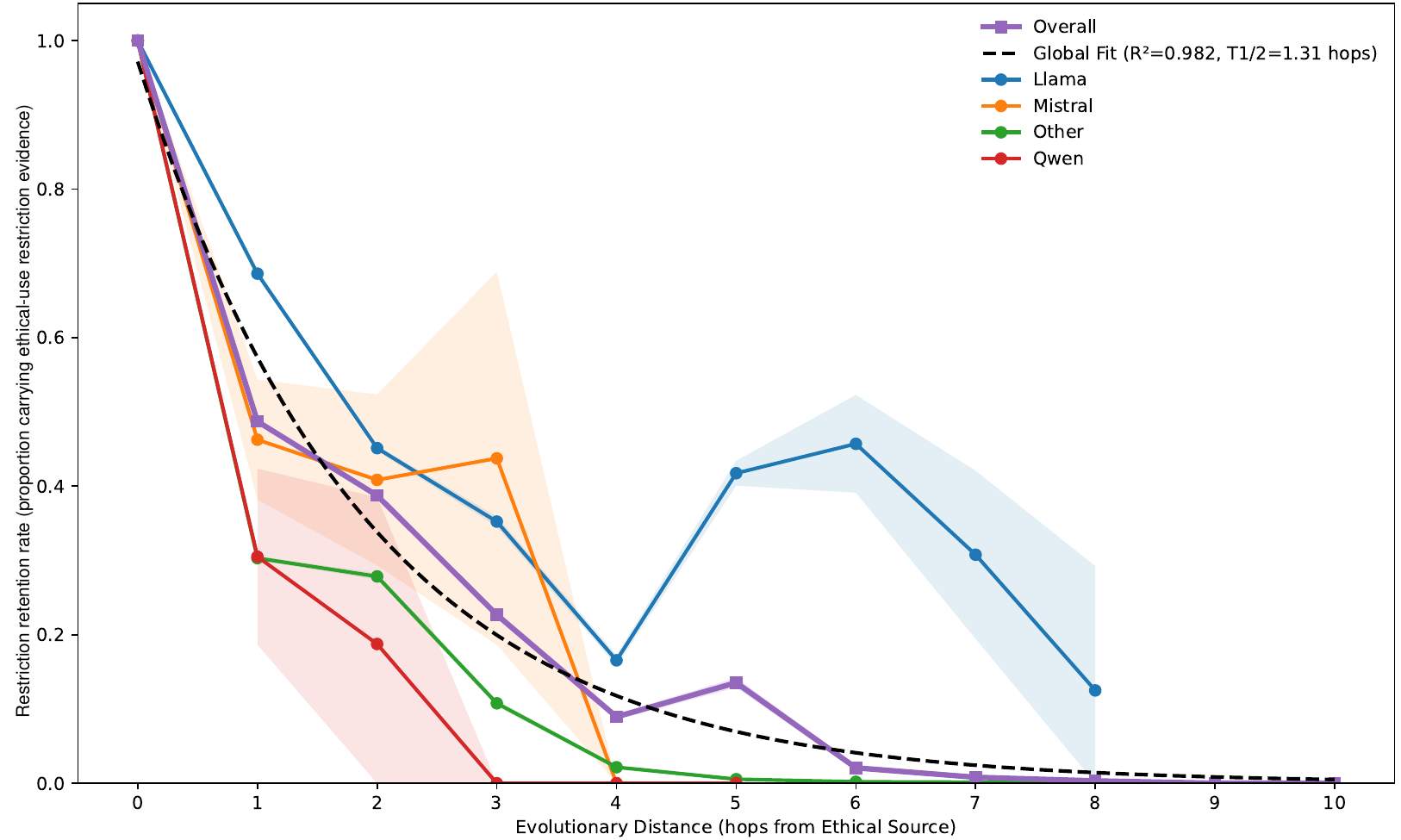}
\caption{\textbf{Ethical-use restriction retention declines with downstream hop distance from ethical source models.} Retention is measured as the proportion of downstream descendants that continue to carry publicly observable ethical-use restriction evidence, using a binary artefact-level classification (see Methods). The purple curve shows overall retention across all validated lineage observations, and coloured curves show major model families (Llama, Mistral, Qwen and others). Shaded bands indicate 95\% normal-approximation confidence intervals ($1.96 \times \mathrm{SEM}$). The dashed line shows an exponential fit to the overall trajectory ($R^2 = 0.982$, $T_{1/2} = 1.31$~hops), indicating rapid dissipation of restriction evidence across downstream generations.}
\label{fig2}
\end{figure}

Across the ecosystem, restriction evidence does not remain stable with lineage depth. Instead, retention declines sharply as downstream hop distance increases (Figure~\ref{fig2}). The overall decay is well approximated by an exponential model, with a fitted half-life of 1.31~hops ($R^2 = 0.982$), indicating that publicly observable ethical-use restriction evidence dissipates rapidly once models begin to propagate downstream. Although retention is mechanically complete at the source level, the fitted trajectory shows a steep decline over the first few generations and approaches zero in deeper downstream lineages.

This pattern is not confined to a single model family. As shown in Figure~\ref{fig2}, major families including Llama, Mistral, Qwen and other lineages display heterogeneous but broadly similar decay profiles, with all exhibiting substantial downstream loss of ethical-use restriction evidence relative to the source level. The first empirical result of our audit is therefore straightforward: in open-weight model genealogies, ethical-use restriction evidence does not reliably inherit across generations, but instead dissipates rapidly with lineage depth.

\subsection{Auditability collapse is closely associated with disclosure failure, but not reducible to missing evidence alone}

To distinguish visible weakening from broader evidentiary collapse, we next classify downstream audit outcomes into four states: Decidable, Inconsistent, Undecidable--Missing (UM) and Undecidable--Ambiguous (UA). Figure~3a shows the full state composition across downstream hop distance from ethical source models. Missing-evidence cases expand rapidly after the source level and become the dominant downstream outcome in deeper generations, indicating that the disappearance of publicly usable evidence is a central feature of lineage propagation.

At the same time, the collapse is not reducible to missing evidence alone. To remove the visual dominance of locally missing cases, Fig.~3b conditions on non-UM cases and examines only the remaining observable or partially observable states. Under this conditional view, Decidable outcomes remain prevalent across much of the graph, but UA becomes increasingly prominent at deeper hops. This shows that even where local evidence is not entirely absent, downstream governance does not simply remain stable. Instead, ambiguity accumulates among the surviving cases, indicating that auditability can deteriorate even within the visible portion of the lineage.

Taken together, these results show that disclosure failure is a major component of governance collapse, but not its whole explanation. Deep lineages do not merely lose metadata; they also contain an increasing share of downstream cases for which the remaining public evidence no longer supports a stable determination.

\begin{figure}[h]
\centering
\includegraphics[width=0.99\textwidth]{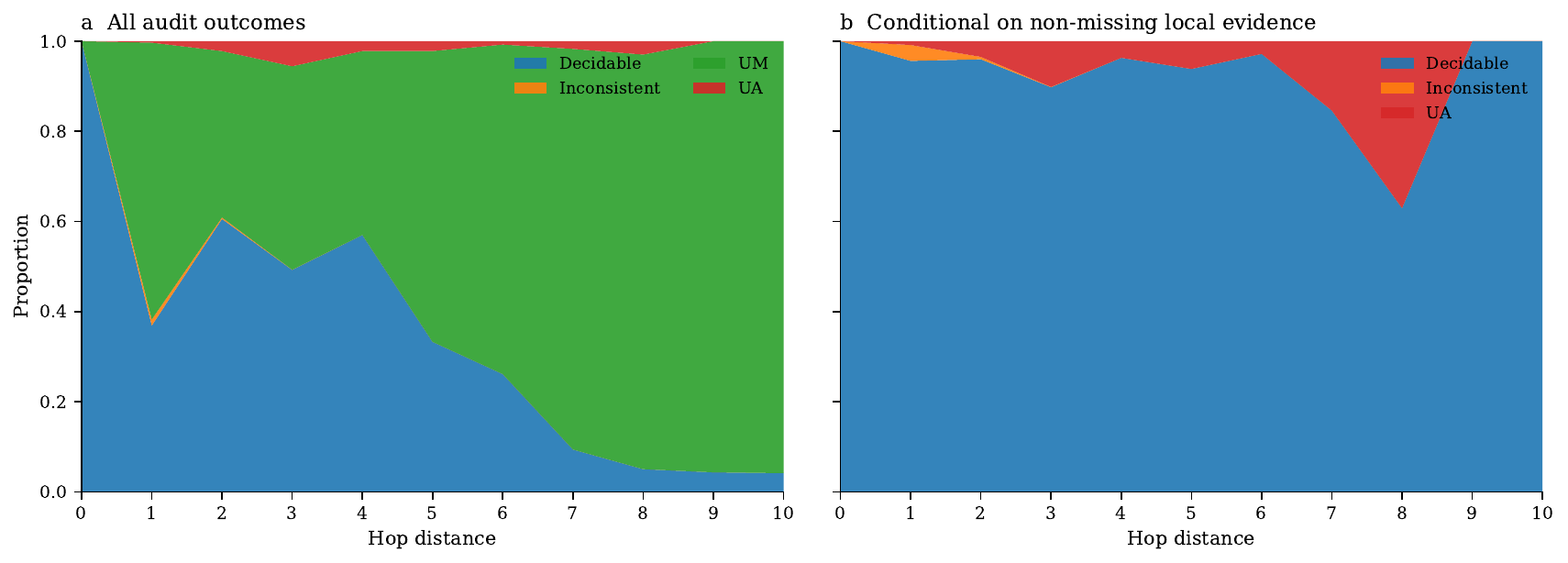}
\caption{Audit outcomes across downstream generations, in the full sample and conditional on non-missing local evidence.
a, Full distribution of downstream audit outcomes by hop, including Decidable, Inconsistent, Undecidable--Missing (UM) and Undecidable--Ambiguous (UA) states. Missing-evidence cases rapidly become the dominant downstream failure mode in deeper generations. b, The same distribution conditional on non-missing local evidence, showing that ambiguity remains and becomes more prominent even when locally missing cases are excluded.}\label{fig3}
\end{figure}

\subsection{Merge conflicts are associated with directional weakening}\label{sec:merge-conflict}

\begin{figure}[t]
\centering
\includegraphics[width=0.60\textwidth]{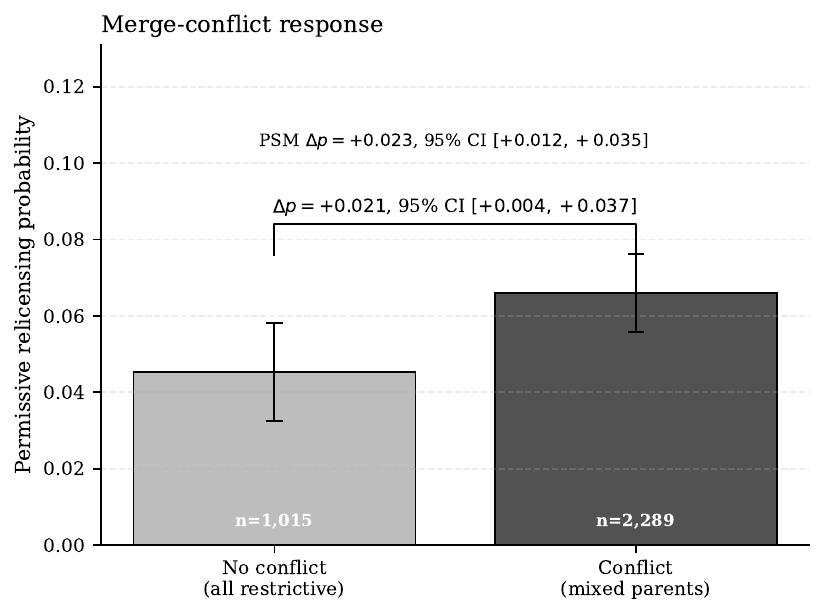}
\caption{\textbf{Mixed-parent merge conflict is associated with higher downstream permissive
relicensing.} The analysis is restricted to cases with at least one direct merge relationship and
at least one scored parent carrying ethical-use restriction evidence. In raw comparisons, permissive relicensing is more frequent under
mixed-parent conflict than under all-restrictive scored parents (0.0660 versus 0.0453; $\Delta p = +0.0206$,
95\% CI $[+0.0039, +0.0367]$, $p = 2.08 \times 10^{-2}$). Under 1:3 propensity-score matching on pre-treatment
covariates with caliper 0.20 and replacement, the estimated difference remains positive ($\Delta p =
+0.0232$, 95\% CI $[+0.0119, +0.0348]$; 2,289 matched treated units). More aggressive robustness analyses (IPW, AIPW) yield directionally similar positive estimates (Supplementary Information Section~S2).}
\label{fig:merge_conflict}
\end{figure}

Mixed-parent merge conflict is associated with higher downstream permissive relicensing across raw and adjusted comparisons. Restricting attention to cases with at least one direct merge relationship and at least one scored parent carrying ethical-use restriction evidence, we find that conflict cases (scored parents include both restrictive and non-restrictive parents) exhibit a higher permissive relicensing probability than no-conflict cases (all scored parents restrictive; 0.0660 versus 0.0453; $\Delta p = +0.0206$, 95\% CI $[+0.0039, +0.0367]$, $p = 2.08 \times 10^{-2}$; Fig.~\ref{fig:merge_conflict}). This raw pattern is consistent with a directional weakening pathway in which mixed upstream constraints are more likely to be followed by downstream permissive relicensing.

We focus on merge because it is the derivation operator that explicitly aggregates multiple upstream sources, creating structurally salient opportunities for licence conflict. Single-parent transformations such as fine-tuning, adapter construction, quantization, distillation and pruning do not by themselves generate mixed-parent governance signals.

Under 1:3 propensity-score matching with caliper 0.20 and replacement on pre-treatment covariates (number of parents, parent missing rate, model age and lineage size; see Methods), the estimated difference remains positive and statistically distinguishable from zero ($\Delta p = +0.0232$, 95\% CI $[+0.0119, +0.0348]$; 2,289 matched treated units). More aggressive quasi-causal robustness analyses on a common-support-trimmed sample yield directionally similar positive estimates: inverse-probability weighting gives an ATE of $+0.0287$ and augmented inverse-probability weighting (doubly robust) gives an ATE of $+0.0294$ (95\% CI $[+0.0085, +0.0484]$; Supplementary Information Section~S2).

We interpret these results with two caveats. First, although balance improves substantially on most covariates after matching, residual imbalance on model age (standardized mean difference $> 0.25$) suggests that unobserved confounders correlated with repository age could partially account for the association. Second, the absolute relicensing probabilities are low in both groups (below 7\%), so the practical magnitude of the effect is modest even though the direction is consistent.

Taken together, these findings suggest that merge remains a governance-critical choke point because it concentrates potentially incompatible upstream signals in a single downstream artefact. The raw association, the matched estimate and the doubly robust estimate all point in the same direction: mixed-parent conflict is associated with a higher probability of downstream permissive relicensing. 

\begin{figure}[h]
\centering
\includegraphics[width=0.62\textwidth]{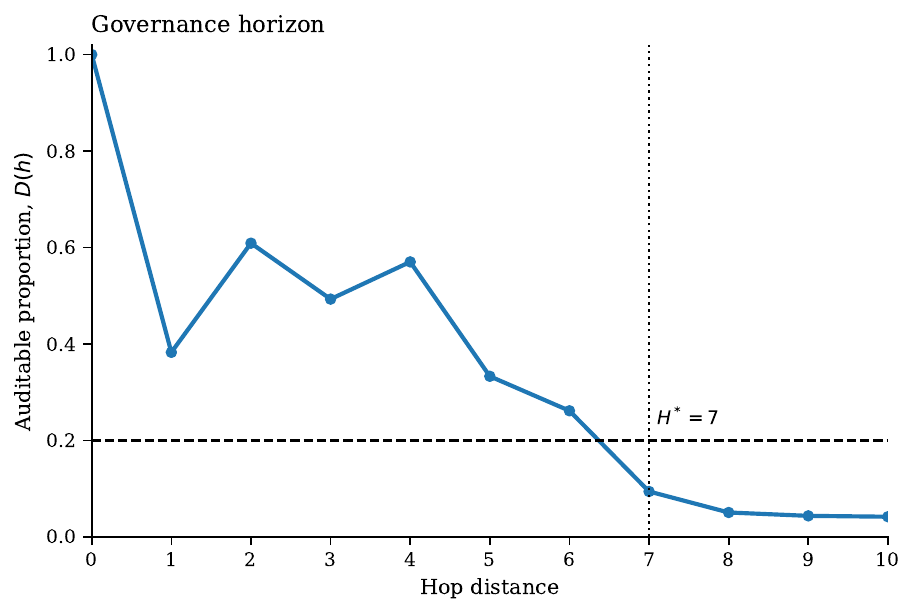}
\caption{\textbf{The auditable proportion reveals an operational governance horizon in open-weight model lineages.} The auditable proportion is defined as $D(h)=1-(UM+UA)$, where $UM$ and $UA$ denote Undecidable--Missing and Undecidable--Ambiguous audit states, respectively. The curve is computed by downstream hop distance from ethical source models at the roots of the validated lineage graph. The dashed line marks the operational threshold $\alpha=0.20$, yielding a governance horizon of $H^*=7$ hops. This estimate is highly stable under bootstrap resampling (mean $=6.98$, 95\% CI $[7.00, 7.00]$).}
\label{fig:governance_horizon}
\end{figure}

\subsection{Open-weight lineages exhibit an operational governance horizon}

Using the auditable proportion $D(h) = 1 - (\mathrm{UM} + \mathrm{UA})$ defined in Methods, we track how the fraction of publicly auditable downstream models changes with hop distance from ethical source models (Figure~\ref{fig:governance_horizon}).

The resulting curve reveals a sharp loss of auditability with lineage depth. Although the auditable proportion is complete at the source level, it falls to 0.38 after one hop, remains only intermittently above 0.5 through hops 2--4, and then declines steeply again, reaching 0.26 at hop~6 and 0.09 at hop~7. Using $\alpha = 0.20$ as the operational threshold, this yields a governance horizon of $H^*(0.20) = 7$~hops. This estimate is highly stable under bootstrap resampling (mean $= 6.98$, 95\% CI $[7.00, 7.00]$; 500~resamples). The governance horizon remains at 7~hops for $\alpha = 0.10$ and at 6~hops for $\alpha = 0.30$, indicating that the result is not sensitive to the choice of operational threshold (Supplementary Section~S3.1).

This result suggests that the practical depth limit of disclosure-based governance in open-weight lineages is shallow. Beyond roughly seven downstream generations, metadata-based auditing is no longer primarily a matter of interpreting visible restrictions; it becomes dominated by missing or ambiguous evidence. Equivalently, once the governance horizon is crossed, at least 80\% of downstream models are no longer publicly auditable under this definition. In this sense, the governance horizon marks the point at which downstream control shifts from a problem of restriction interpretation to a problem of evidentiary collapse.

\begin{figure}[h]
\centering
\includegraphics[width=0.72\textwidth]{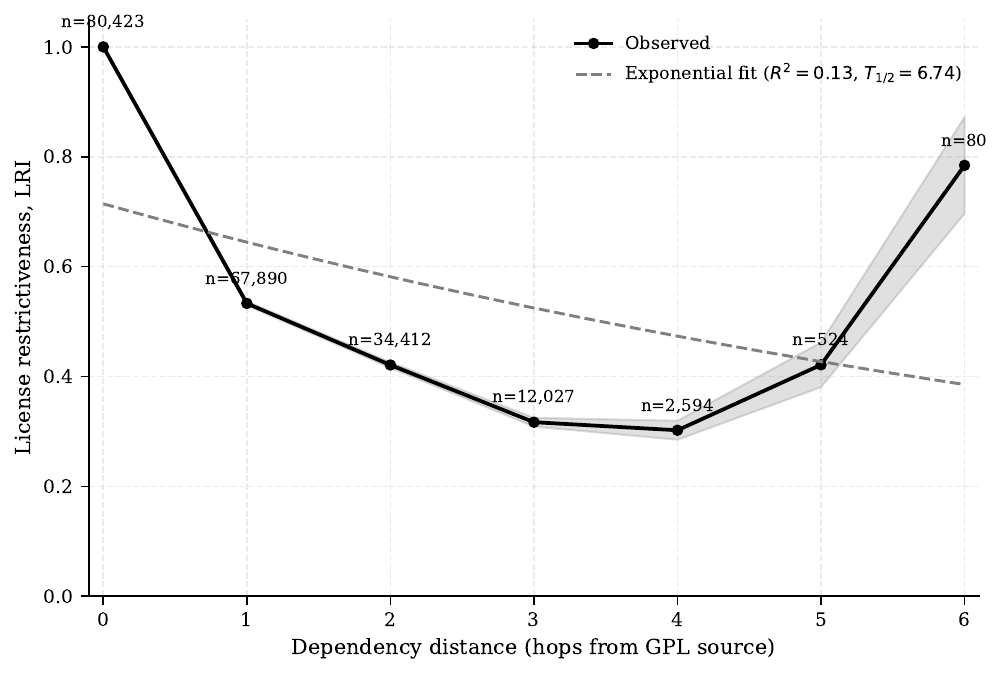}
\caption{\textbf{PyPI software-dependency comparator does not exhibit the same depth-bounded collapse as open-weight model lineages.} The curve shows mean downstream license restrictiveness from strict GPL source packages, measured using a license restrictiveness index (LRI). Shaded bands indicate 95\% confidence intervals, and sample sizes are shown at each hop. Although restrictiveness declines at early dependency distances, the trajectory does not collapse monotonically with depth and is only weakly approximated by an exponential fit ($R^2 = 0.13$, $T_{1/2}=6.74$ hops). This comparator therefore contrasts with the sharp downstream auditability collapse observed in open-weight model lineages.}
\label{fig:pypi_control}
\end{figure}

\subsection{The collapse is topology-specific to open-weight derivation}

To assess whether deep governance collapse is a generic feature of open ecosystems, we compare open-weight model lineages with a software-dependency comparator ecosystem based on PyPI. Specifically, we track downstream licence restrictiveness from strict GPL source packages using a licence restrictiveness index (LRI), which maps each package's licence to a $[0, 1]$ scale where 1.0 corresponds to strong copyleft (GPL) and 0.0 to fully permissive terms (see Methods). The comparator is not intended to equate models with software packages in a legal or technical sense. Rather, it provides a contrasting evolutionary topology against which to evaluate whether the sharp depth-bounded auditability collapse observed in Hugging Face is specific to open-weight derivation.

Figure~\ref{fig:pypi_control} shows that the PyPI comparator does exhibit some early decline in downstream restrictiveness, but it does not display the same monotonic collapse seen in open-weight model lineages. Mean restrictiveness drops from 1.00 at the source level to 0.53 at one hop and 0.42 at two hops, but then stabilizes rather than continuing toward zero, remaining at 0.30--0.42 through hops 3--5. The apparent upturn at hop~6 is based on only 80~observations and is not statistically distinguishable from the hop~3--5 plateau given the wide confidence interval at that depth.

The contrast with open-weight lineages is sharp. In the Hugging Face ecosystem, restriction evidence decays exponentially with a half-life of 1.31~hops ($R^2 = 0.98$); while in PyPI, the exponential fit is weak ($R^2 = 0.13$, $T_{1/2} = 6.74$~hops) and downstream restrictiveness stabilizes rather than collapsing toward zero. This comparison suggests that the governance horizon identified in open-weight AI is not a generic consequence of openness itself. Rather, it reflects the specific sociotechnical conditions of weight-level reuse, where governance signals depend on repeated voluntary restatement rather than on strongly inherited dependency artefacts.

\section{Discussion}

\subsection{Implications for AI governance}

Our findings imply that current regulatory expectations of supply-chain traceability for open-weight AI cannot be met through voluntary disclosure alone beyond shallow lineage depths. The EU AI Act imposes transparency and supply-chain documentation obligations on providers and deployers of high-risk and general-purpose AI systems~\cite{euaiact2024,veale2021demystifying}; recent U.S.\ policy reviews have examined whether disclosure and licensing mechanisms are adequate for dual-use foundation models with widely available weights~\cite{ntia2024dualuse}, and the U.S.\ NIST AI Risk Management Framework calls for documentation and provenance practices that span the full AI supply chain~\cite{ai2023artificial,cen2023supply}; and the OECD has highlighted that open-weight model governance requires context-specific policy calibration, noting that open-weight models now account for approximately 55\% of commercially available foundation models worldwide as of April~2025~\cite{oecd2025aiopenness}. Each of these frameworks implicitly assumes that disclosed restrictions remain visible as models propagate downstream. The empirical governance horizon of seven derivation hops indicates that this assumption is structurally unsupported in current practice, and that compliance regimes calibrated against a few generations of lineage will encounter populations of artefacts for which the necessary public evidence has already collapsed.

\begin{figure}[t]
\centering
\includegraphics[width=0.75\textwidth]{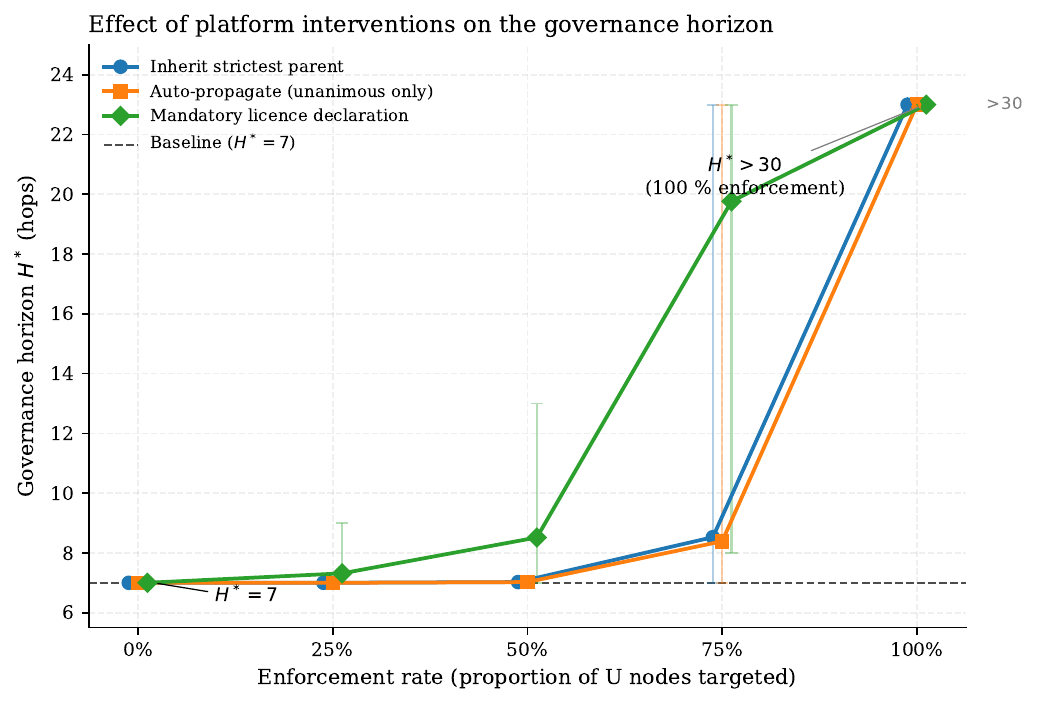}
\caption{\textbf{Effect of platform interventions on the governance horizon.} Three counterfactual designs are simulated for resolving Undecidable--Missing (U) nodes (Supplementary Information Section~S4). The enforcement rate denotes the proportion of U nodes targeted by the intervention. \emph{Inherit-strictest} and \emph{auto-propagate} resolve a targeted U node only when at least one ancestor already carries a Permissive or Restrictive intent; they behave nearly identically. Below 50\% enforcement, the governance horizon does not move ($H^* = 7$). At 75\% enforcement, Monte Carlo simulation intervals widen sharply ([7,\,$>$30]), indicating a transitional regime sensitive to which nodes are resolved. Only at near-complete enforcement does $H^*$ exceed 30~hops. \emph{Mandatory licence declaration} additionally forces a definite intent on orphan U nodes (defaulting to Restrictive when no inheritable upstream exists), and yields a substantively different trajectory: the horizon already moves at 25\% enforcement (mean $H^* = 7.3$) and rises to mean $H^* = 19.8$ at 75\%. The contrast identifies the structural bottleneck as the existence of lineage components without inheritable upstream intent, rather than the strength of the inheritance rule alone. Upward arrows denote right-censored estimates ($H^* > 30$). Error bars show 95\% Monte Carlo simulation intervals over 500 stochastic enforcement realizations.}
\label{fig:intervention}
\end{figure}

Mandatory disclosure has a bounded reach, and the design of the policy itself matters. Our intervention simulation (Figure~\ref{fig:intervention}; Supplementary Information Section~S4) compares three platform-level designs: inherit-strictest parent, unanimous auto-propagation, and mandatory licence declaration. The first two designs resolve a targeted Undecidable--Missing (U) node only when at least one of its ancestors already carries a Permissive or Restrictive intent; they behave nearly identically. Below 50\% enforcement the governance horizon does not move at all; at 75\% the system enters a transitional regime with wide Monte Carlo simulation intervals ($H^*$ ranging from 7 to $>$30~hops depending on which subset of nodes is resolved); only near-complete enforcement extends auditability to deep lineages. The explanation is structural and operates through two mechanisms. First, an unresolved upstream node propagates as an undecidable and ambiguous state on its immediate downstream children, creating a local bottleneck on each path through it. Second, more consequentially for inheritance-only policies, many U nodes belong to lineage components with no inheritable P/R ancestor anywhere upstream; these orphan components remain unaddressable under any inheritance-only intervention, regardless of enforcement rate. The third design, mandatory licence declaration, additionally requires every targeted U node to receive a definite intent (defaulting to restrictive when no inheritable upstream exists), and this single change relaxes the bottleneck: the horizon already moves at 25\% enforcement (mean $H^* = 7.3$, MC interval [7,\,9]), reaches mean $H^* = 8.5$ at 50\% (MC interval [7,\,13]), and rises to mean $H^* = 19.8$ at 75\% (MC interval [8,\,$>$30]). Achieving deep auditability through platform-level intervention therefore requires not only high enforcement but also a design that resolves orphan-U lineage components, not just an inheritance rule layered on top of voluntary disclosure. Either condition is hard to meet in practice across more than two million repositories with heterogeneous documentation practices~\cite{gorwa2024moderating}.

The policy implication is that disclosure-based governance has a structurally bounded reach in open-weight AI, and that achieving deep supply-chain accountability will require provenance-preserving infrastructure that carries governance signals through derivation itself. Concrete directions include cryptographic provenance attestation, machine-readable licence chains embedded in model weights, and platform-enforced derivation registries that record inherited restriction lineages independently of downstream disclosure choices~\cite{bateman2024beyond,bommasani2023ecosystem}; analogous proposals have been advanced in policy comments to copyright authorities urging mandatory data and model provenance~\cite{mahari2023comment}. We do not view this as a substitute for disclosure: mandatory licence declaration, in particular, would substantially reduce undecidability in the first few hops where the largest absolute number of models reside. Rather, our results suggest that disclosure and provenance attestation are complementary, with the latter required specifically to extend governance reach beyond the empirical horizon identified here. Policymakers calibrating traceability requirements to open-weight ecosystems may therefore need to consider whether mandatory provenance-preserving mechanisms, rather than voluntary artefact-level disclosure, are necessary to support deep supply-chain accountability.

\subsection{Why open-weight derivation differs in different ecosystem}

In open-weight AI, technical capability propagates through weight-level derivation while governance signals propagate, if at all, through metadata. This asymmetry is not a documentation shortcoming but a structural mismatch between the expected reach of restrictions and the mechanisms available to carry them. When a model is fine-tuned, merged, or quantized, the weights are transformed, but no governance information is automatically inherited; whether downstream restrictions remain visible depends entirely on whether successive developers choose to restate them. The failure mode that follows is not directly observable noncompliance but \emph{undecidability}: as lineage depth increases, missing and ambiguous evidence rapidly dominate downstream audit outcomes, shifting the practical question from how to interpret a restriction to whether sufficient public evidence remains to determine that a restriction exists at all. We emphasise that this horizon is a claim about public auditability, not about the ontological status of upstream obligations: legal or policy constraints do not cease to exist after a fixed number of derivation steps, but the practical ability to recover, inspect, and attribute such constraints from public artefacts alone does.

The PyPI comparator isolates the source of this collapse. Software dependencies are mediated through explicit, machine-readable reference relations that remain inspectable at considerable depth, whereas open-weight derivation proceeds through weight-level transformation and depends on repeated voluntary restatement of governance signals. What propagates most reliably across open-weight derivation is therefore capability, not governance. The central governance challenge in open-weight AI is not openness itself but the topology where reuse occurs, which intersects with broader debates on the gradient of AI openness~\cite{solaiman2023gradient,osi2024osaid}, frameworks for classifying model openness along multiple dimensions~\cite{white2024model}, the downstream risks of open-sourcing capable foundation models~\cite{seger2023opensourcing}, the political economy of who actually benefits from ``open'' AI infrastructure~\cite{widder2023open}, and parallel legal analyses of when AI training and downstream model derivatives qualify as fair use or constitute new derivative works~\cite{henderson2023foundation,lemley2020fair,gervais2021ai}.

This positioning sharpens our contribution relative to prior ecosystem audits. Where earlier work has catalogued attribution failures in linear dataset pipelines~\cite{longpre2024large}, incomplete artefact-level disclosure in model and dataset documentation across a series of complementary documentation frameworks~\cite{mitchell2019model,gebru2021datasheets,bender2018data,pushkarna2022data,bommasani2023foundation,pepe2024hugging,yangnavigating}, holistic evaluation of foundation models in shared registries~\cite{liang2022holistic}, widespread licensing conflict and noncompliance in AI supply chains~\cite{duan2025position,jewitt2026permissive,wang2026hidden,jewitt2025hugging}, and the legal structure of the generative-AI supply chain~\cite{lee2023talkin}, our analysis identifies depth-bounded undecidability as the dominant failure mode in topologically richer model lineages. Concurrent work has observed that licences in the Hugging Face ecosystem drift directionally toward more permissive terms within model family trees~\cite{oderinwale2025anatomy}. We extend this observation by quantifying merge-specific upstream conflict as a governance choke point with quasi-causal robustness checks and by formalizing the depth dependence of disclosure failure as an operational horizon.

\subsection{Limitations and outlook}

Three limitations bear on the interpretation of our findings. First, our analysis is based on a single-platform snapshot --- the Hugging Face model ecosystem in October 2025 --- and does not directly establish whether the governance horizon is contracting, stable, or expanding over time, nor whether comparable horizons exist in other open-weight model registries; we therefore release a deterministic analysis pipeline so that subsequent cohorts and platforms can be benchmarked against the same protocol~\cite{pineau2021improving}. Second, our restriction signal is derived from publicly disclosed licence text and abstracts away from the legal force of those disclosures: a restriction visible in metadata may not remain enforceable in practice, and obligations not appearing in disclosures may persist as a matter of law. The governance horizon delimits the reach of disclosure-based governance \emph{infrastructure} rather than the reach of law itself. Third, although the merge-conflict analysis identifies a positive association between mixed-parent merge and downstream permissive relicensing across raw, matched, and doubly robust estimators, residual imbalance on model age after matching means that unobserved confounders correlated with repository age cannot be fully excluded; definitive causal attribution would require richer covariate adjustments or quasi-experimental variation in merge composition.

Our results identify provenance attestation, rather than disclosure standardisation, as the binding technical requirement for deep governance reach in open-weight AI. Promising directions include cryptographic licence-chain attestation, weight-embedded provenance signatures, and platform-enforced derivation registries that record inherited restriction lineages independently of downstream disclosure choices. Such mechanisms remain technically nascent and will require coordinated platform, regulatory, and research effort to deploy at ecosystem scale. Establishing whether they can in practice extend the governance horizon --- and at what cost to ecosystem openness --- is the natural next empirical question.

\section{Methods}
\label{Methods}
\subsection{Data sources}

The empirical setting is the Hugging Face model ecosystem~\cite{wolf2020transformers}. Our October 2025 snapshot was collected by exhaustively enumerating all public model repositories through the Hugging Face Hub API (\texttt{list\_models}, full metadata mode), yielding 2{,}142{,}823 model artefacts. For each repository we record author, creation and modification timestamps, pipeline tags, library identifier, 30-day and all-time download counts, likes and tag set, and retrieve the full model card (README.md), parsing both the YAML front-matter and the free-text body. Retrieval runs in parallel with exponential back-off; access-restricted repositories are excluded.

Licence information is collected at three levels: (i) repository-level tags of the form \texttt{license:<name>}, (ii) the \texttt{license} field of model-card YAML, and (iii) the full licence-file text, downloaded across 14 common filename variants (\texttt{LICENSE}, \texttt{LICENSE.md}, \texttt{LICENSE.txt}, \texttt{COPYING}, \texttt{NOTICE}, and their case variants). The mapping from these signals to ethical-use restriction labels is described in Section~\ref{sec:restriction-classification}.

From this snapshot we extract 1{,}033{,}781 validated model-to-model derivation relationships (Section~\ref{method_dep_extract}); dataset edges are excluded throughout. The PyPI comparator ecosystem is described in Section~\ref{sec:pypi-control}.

\subsection{Dependency extraction and validation}
\label{method_dep_extract}

We extract candidate parent–child derivation relationships from three layers of repository evidence and apply rule-based noise filtering before edge typing.

\textbf{Structured metadata.} YAML front-matter fields (\texttt{base\_model}, \texttt{base\_models}, \texttt{datasets}) and repository-level tags (e.g.\ \texttt{base\_model:finetune:org/repo}) are parsed first and retained unconditionally.

\textbf{Textual model cards.} The README extractor handles three contexts: inline hyperlinks and explicit prose (``fine-tuned from [org/model]''); markdown tables, separating derivation-relevant from benchmark tables via column-header detection (``precision'', ``F1'', ``MMLU''); and repository-name patterns indicative of derivation (quantization suffixes \texttt{-GGUF}, \texttt{-AWQ}, \texttt{-GPTQ}).

\textbf{Entity resolution.} Extracted references are resolved against the full 2{,}142{,}823 model-ID set by exact match, then by fuzzy match using official organization prefixes (e.g.\ \texttt{meta-llama}, \texttt{mistralai}, \texttt{Qwen}) and suffix normalization. Unresolvable references are discarded.

\textbf{Edge typing.} Each retained edge is assigned a derivation type from a fixed taxonomy: fine-tuning, adapter, merge, quantization, distillation, pruning, or \texttt{base\_model} (default when a parent–child link is supported but no more specific operator can be assigned). Tag- and YAML-derived signals take precedence over name-pattern heuristics and textual cues. When multiple types are assigned to the same pair, conflict resolution favours the most specific operator: quantization $>$ adapter $>$ finetune $>$ merge $>$ \texttt{base\_model}.

\textbf{Noise filtering.} Benchmark tables are excluded; within non-benchmark tables only rows naming the current repository are processed (self-row targeting); quantization models with multiple candidate parents retain only the highest string-similarity match; sibling or same-family models in list contexts are removed unless corroborated by metadata.

\textbf{Evaluation.} We benchmark the pipeline against an LLM judge (GPT-5) on a stratified sample of 417 repositories: standard derivation types ($n = 197$); rare types --- distillation, pruning, quantization, merge ($n = 110$); and true-negative repositories ($n = 110$). Population metrics are obtained by Horvitz–Thompson expansion. Under topology-level evaluation (ID-only match), micro-$F_1 = 94.0\%$ (precision 94.2\%, recall 93.8\%; true-negative rate 68.9\%; sample-level accuracy 96.3\%). Strict evaluation (ID + exact type) yields micro-$F_1 = 84.6\%$ (precision 84.7\%, recall 84.4\%; sample accuracy 93.1\%); type-compatible evaluation, which allows specific operators to be coarsened to \texttt{base\_model}, yields micro-$F_1 = 85.1\%$.

To assess judge stability, a stratified subsample of 60 repositories (S1 = 28, S2 = 16, S3 = 16) was independently re-annotated by a second model (GPT-5.4) under the same prompt, blind to the original outputs. After excluding dataset edges, exact-match agreement over complete model-to-model dependency sets was 0.883 (Cohen's $\kappa = 0.865$); edge-level precision, recall and $F_1$ were 0.886, 0.848 and 0.867. The seven disagreement cases were manually categorised against canonical Hugging Face tag conventions (Supplementary Table~S2); most reflect dependency-type boundary cases rather than different parent structures.

\subsection{Lineage graph construction}
\label{sec:graph-construction}

Validated edges from Section~\ref{method_dep_extract} are assembled into the ecosystem-wide lineage graph used in all subsequent analyses. Retained edge types include fine-tuning, adapter, merge, quantization, distillation, pruning, conversion and \texttt{base\_model}; dataset edges are excluded.

For each model, we compute the full dependency subgraph by breadth-first traversal from the model as root, following parent edges and recording all reachable edges. Cycles (e.g.\ mutual adapter references) are handled with a per-path visit set; a safety cutoff of 30~hops and 500{,}000 traversal steps is imposed per root.

Within each subgraph we record \emph{lineage depth} as the longest-path distance from root to each reachable ancestor (not shortest), so that depth reflects a model's deepest derivation chain --- the relevant quantity when a restriction must survive every chain from source. All traversed edges are retained regardless of whether they lie on the longest path.

The resulting graph contains 1{,}033{,}781 validated model-to-model edges. For the hop-indexed analyses, source models are lineage roots (in-degree zero) carrying ethical-use restriction evidence (Section~\ref{sec:restriction-classification}); downstream hop distance is shortest-path distance from the nearest such source, with a cutoff of 10~hops. This yields 1{,}916 ethical source models and 133{,}812 reachable descendants within the analysis window.

\subsection{Ethical-use restriction classification}
\label{sec:restriction-classification}

We classify each repository's licence evidence as carrying publicly observable ethical-use restriction signals using a rule-based heuristic over full licence text. When a repository's own licence file is available, we use that text directly; otherwise, if the repository carries a recognized licence name in its tags or YAML, we substitute the canonical template text for that licence. Repositories with neither retrievable licence text nor a recognized identifier are coded as missing evidence and carried forward as a distinct category in the audit-state analysis (Section~\ref{sec:audit-state}).

The classifier scores three signal layers additively: (i) prohibition framing (``must not'', ``shall not'', ``prohibited'', ``not permitted'') together with explicit policy references (``acceptable use policy'', ``responsible use''); (ii) harm-domain terms within a restriction context, with higher weights on military, surveillance, child sexual exploitation, hate speech, malware, election interference, high-risk decision-making, medical, legal, and financial domains, and lower weights on misinformation, harassment, privacy, fraud, and deceptive use; (iii) explicit references to known responsible-use licence families (RAIL, OpenRAIL, Llama Community License) when co-occurring with prohibition framing or policy language. Harm-domain matches are counted only when they occur within a restriction context --- co-occurrence with prohibition framing in the same sentence or proximity to a restriction-related section heading within a fixed window --- which suppresses false positives from generic legal-compliance text that lists harms without imposing behavioural constraints. A repository is classified as carrying ethical-use restriction evidence when the aggregate score exceeds 1.0. We focus on licence text because ethical-use restrictions, when formalized, are most consistently encoded at the licence layer.

\textbf{Evaluation.} The classifier is evaluated against a gold standard of 200 licence texts, sampled by a hybrid frequency strategy: standard licences are aggregated by name, custom or unspecified licences by text content, and the resulting groups ranked by ecosystem frequency. Each licence was independently coded by an LLM and a domain-informed human annotator; the labels agreed on 191 of 200 (agreement 0.955; Cohen's $\kappa = 0.908$). The nine disagreements were resolved by discussion-based adjudication, under a protocol that counts behavioural restrictions incorporated through mandatory acceptable-use policies as ethical-use restriction evidence; seven retained the LLM label and two the human label. On the adjudicated gold set, the classifier achieves precision $= 0.96$, recall $= 0.91$, $F_1 = 0.94$ and accuracy $= 0.95$ (TP = 81, FN = 8, FP = 3, TN = 108). The pattern inventory, scoring scheme, codebook, sampling procedure and adjudication results are in Supplementary Information Section~S6.

\subsection{Audit-state definitions}
\label{sec:audit-state}

To analyse governance as a problem of auditability rather than directly observable violation, each downstream model is assigned one of four audit states from the publicly available licence evidence at the model and its upstream ancestors.

\textbf{Intent assignment.} Each node receives a licence intent: \emph{restrictive}~(R) if the classifier (Section~\ref{sec:restriction-classification}) flags ethical-use restriction evidence; \emph{permissive}~(P) if its licence name matches a recognized permissive SPDX identifier (e.g.\ MIT, Apache-2.0, BSD-3-Clause); \emph{unknown}~(U) otherwise.

\textbf{State assignment.} Because the lineage graph may contain cycles, we first condense it into a directed acyclic graph by collapsing strongly connected components; intent and merge status are aggregated at component level. States are then assigned in topological order under the following priority:

\begin{enumerate}[leftmargin=*, nosep]

\item \textbf{Undecidable--Missing (UM).} The node's own intent is U: local evidence required for any governance determination is absent.

\item \textbf{Undecidable--Ambiguous (UA) via merge conflict.} The node is a merge product (merge tags, multiple parents, or merge-specific YAML) with mixed R and P upstream intents, and the available merge-level evidence falls below a predefined threshold: conflicting signals cannot be reconciled from public artefacts alone.

\item \textbf{UA via upstream missingness.} The node's own intent is not U but at least one ancestor is UM: governance status depends on upstream evidence that is itself unavailable.

\item \textbf{Inconsistent.} The node declares a permissive intent but inherits from an upstream node carrying a passthrough-restrictive licence (specifically, an OpenRAIL-family licence whose terms propagate to derivatives).

\item \textbf{Decidable.} None of the above applies: the available evidence is sufficient to determine the applicable restriction status.

\end{enumerate}

The main analysis uses a strict reconciliation policy for resolved merge conflicts and treats upstream missingness as ambiguous rather than missing. Sensitivity to these choices is examined in Supplementary Information Section~S3.2.

\subsection{Governance horizon formalization}
\label{sec:governance-horizon}

The \emph{auditable proportion} at hop~$h$ is
\[
D(h) \;=\; 1 - \bigl(\mathrm{UM}(h) + \mathrm{UA}(h)\bigr),
\]
where $\mathrm{UM}(h)$ and $\mathrm{UA}(h)$ are the proportions of downstream models in Undecidable--Missing and Undecidable--Ambiguous states at hop~$h$ (Section~\ref{sec:audit-state}). $D(h)$ thus measures the fraction of downstream models for which the available public evidence supports a governance determination (Decidable or Inconsistent).

The \emph{operational governance horizon} is the smallest hop distance at which $D(h)$ falls to or below threshold~$\alpha$:
\[
H^*(\alpha) \;=\; \min\bigl\{\,h : D(h) \leq \alpha\,\bigr\}.
\]
We use $\alpha = 0.20$ as the primary threshold: beyond $H^*$ at least 80\% of downstream models are no longer publicly auditable under this definition. When $D(h)$ does not reach $\alpha$ within the observation window, $H^*$ is right-censored at $\text{max\_hop} + 1$.

Source models, hop distance and the analysis window are defined as in Section~\ref{sec:graph-construction}: 1{,}916 ethical lineage roots and 133{,}812 reachable descendants within 10~hops. Stability of $H^*$ is assessed by bootstrap resampling the node-level audit-state observations 500~times with replacement; we report the bootstrap mean and 95\% percentile CI. Robustness is examined along three dimensions: $\alpha \in \{0.10, 0.20, 0.30, 0.40\}$; the merge-evidence threshold $\tau \in \{1, 2, 3, 4\}$; and the reconciliation policy (strict, lenient, none) crossed with the upstream-missingness propagation policy (ambiguous, missing), giving $4 \times 3 \times 2 = 24$ combinations. Full sensitivity results are in Supplementary Information Sections~S3.1 and~S3.2.
\subsection{Statistical analyses}
\label{sec:statistical-analyses}

\textbf{Restriction retention.} Retention at hop~$h$ is the proportion of downstream descendants of ethical source models that continue to carry publicly observable ethical-use restriction evidence (Section~\ref{sec:restriction-retention}). The unit of observation here is a (source, descendant) lineage observation taken across all validated lineage paths in the ecosystem and aggregated by shortest hop distance to any ethical source; this differs from the unique-node analysis window of $1{,}916$ global ethical roots and $133{,}812$ unique reachable descendants used for the audit-state and governance-horizon analyses (Section~\ref{sec:graph-construction}). We fit an exponential decay model to the overall trajectory and report the fitted half-life~$T_{1/2}$ and~$R^2$; family-specific curves for Llama, Mistral, Qwen, and others use 95\% normal-approximation CIs ($1.96 \times \mathrm{SEM}$) at each hop.

\textbf{Merge-conflict analysis.} For Section~\ref{sec:merge-conflict} we restrict to cases whose direct dependencies include at least one merge edge and in which at least one scored parent carries ethical-use restriction evidence (score $\geq 1.0$). \emph{Conflict} cases have scored parents of mixed restrictive/non-restrictive intent; \emph{no-conflict} cases have all-restrictive scored parents. Parents with missing scores enter the propensity model through the parent missing-rate covariate but do not define treatment status. The outcome is binary permissive relicensing (child carries a recognized permissive licence name); children with missing licence information are conservatively coded as non-relicensed.

We report four estimators of increasing identification strength: (i)~the raw proportion difference with a two-sided $z$-test and bootstrap 95\% CI; (ii)~propensity-score matching~\cite{rosenbaum1983central} (scaled logistic regression on number of parents, parent missing rate, log model age, log lineage size, their squares, and one interaction; 1:3 nearest-neighbour, caliper 0.20, with replacement; ATT with clustered bootstrap CI); (iii)~inverse-probability weighting on the common-support-trimmed sample ($e \in [0.05, 0.95]$, propensity refit after trimming); and (iv)~augmented inverse-probability weighting~\cite{robins1994estimation} combining the propensity model with separate outcome regressions for treated and control on the trimmed sample. Bootstrap CIs for IPW and AIPW use 800 resamples, refitting all relevant models on each resample. Covariate balance is assessed by standardized mean differences before and after matching. Full results, balance diagnostics and propensity-score overlap plots are in Supplementary Information Section~S2; the main text reports the raw and PSM estimates, with IPW and AIPW as additional robustness checks.

\subsection{PyPI comparator ecosystem}
\label{sec:pypi-control}

To test whether deep governance collapse is specific to open-weight lineages or generic to open ecosystems, we build a software-dependency comparator on the Python Package Index (PyPI). PyPI is structurally distinct: dependencies are declared through machine-readable manifests rather than restated through voluntary textual disclosure. The comparison is not designed to equate models with software packages legally or technically; rather, it provides a contrasting evolutionary topology against which to evaluate whether the depth-bounded auditability collapse observed in open-weight lineages is topology-specific.

We use a snapshot of PyPI release metadata (names, versions, declared dependencies, licence fields). For each release we extract the dependency tree from structured metadata and build a directed graph in which edges point from upstream dependencies to downstream dependants. To prevent frequently updated packages from dominating the hop distribution, we apply time-stratified sampling: at most one release per package per calendar year, selected by deterministic hashing.

Each package's licence is mapped to a \emph{licence restrictiveness index} (LRI) on $[0, 1]$ using a reference table of 63 licence terms graded by copyleft strength: strong copyleft (GPL-3.0, AGPL-3.0) at 1.0; weak copyleft (LGPL, MPL) at 0.75 or 0.5; permissive (MIT, Apache-2.0, BSD) at 0.0. Licence strings are matched by exact lookup on the normalised name, then by substring matching for non-standard declarations. Source packages are dependency roots (in-degree zero) with LRI\,=\,1.0 (GPL-family); downstream dependency distance is the shortest-path distance from the nearest GPL source, with a cutoff of 6~hops. We fit the same exponential decay model used in the Hugging Face analysis. The complete LRI mapping, sampling procedure and per-hop statistics are in Supplementary Information Section~S5.

\bibliography{sn-bibliography}

\setcounter{section}{0}
\renewcommand{\thesection}{S\arabic{section}}
\renewcommand{\thesubsection}{S\arabic{section}.\arabic{subsection}}
\renewcommand{\thefigure}{S\arabic{figure}}
\renewcommand{\thetable}{S\arabic{table}}
\setcounter{figure}{0}
\setcounter{table}{0}

\section{Dependency extraction and annotation validation}
\label{sec:supp_dependency_validation}

This section provides additional details on the validation of the dependency-extraction pipeline described in Methods Section~4.2. We first evaluate the extraction pipeline against judge-derived reference annotations on a stratified sample of 417 repositories, then assess the stability of the LLM-based reference annotations through an independent blind re-annotation of a 60-repository subsample.

\subsection{Stratified evaluation against an LLM judge}

The sampling unit is a Hugging Face repository (not an edge), because the extraction task is to recover the complete set of direct parent dependencies per repository. The evaluation sample contains 417 repositories across three strata: \textbf{S1 Standard} ($n = 197$), with common dependency types; \textbf{S2 Rare} ($n = 110$), oversampled for governance-relevant rare operators (distillation, pruning, quantization, merge); and \textbf{S3 True negative} ($n = 110$), repositories whose judge-derived direct dependency set is empty.

For each sampled repository, the pipeline output is compared against a reference annotation produced by prompting GPT-5 with the repository identifier, tags, YAML metadata and full model card. The judge extracts only direct parent dependencies (not grandparents) and assigns each to one of eight relationship types (\texttt{base\_model}, \texttt{finetune}, \texttt{adapter}, \texttt{quantization}, \texttt{merge}, \texttt{distillation}, \texttt{pruning}, \texttt{dataset}); the protocol prioritises structured Hugging Face metadata over README prose and ignores benchmark or leaderboard tables.

We evaluate under three modes: \textbf{topology-level} (parent ID match only), \textbf{strict} (parent ID + exact relationship type), and \textbf{type-compatible} (parent ID match, with specific types allowed to coarsen to \texttt{base\_model}). Population-level metrics are obtained by Horvitz--Thompson expansion: with $N_s$ the estimated population size of stratum $s$ and $n_s$ the sampled size, each sampled repository receives weight $w_s = N_s / n_s$; weighted TP/FP/FN are aggregated across strata before computing micro-averaged precision, recall and $F_1$ (Table~\ref{tab:s1_pipeline_validation}). Topology-level evaluation yields micro-$F_1 = 94.0\%$ (precision 94.2\%, recall 93.8\%) with true-negative rate 68.9\% and sample-level accuracy 96.3\%; strict evaluation yields micro-$F_1 = 84.6\%$ (precision 84.7\%, recall 84.4\%, accuracy 93.1\%); type-compatible evaluation yields micro-$F_1 = 85.1\%$.

\begin{table}[h]
\centering
\caption{Pipeline evaluation against GPT-5 judge-derived reference annotations.}
\label{tab:s1_pipeline_validation}
\begin{tabular}{lccccc}
\hline
Evaluation mode & Precision & Recall & Micro-$F_1$ & Accuracy & TNR \\
\hline
Topology-level & 0.942 & 0.938 & 0.940 & 0.963 & 0.689 \\
Strict typed & 0.847 & 0.844 & 0.846 & 0.931 & -- \\
Type-compatible & -- & -- & 0.851 & -- & -- \\
\hline
\end{tabular}
\end{table}

\subsection{Independent re-annotation stability}

To assess judge stability we draw a stratified subsample of 60 repositories from the evaluation set (S1 = 28, S2 = 16, S3 = 16) and re-annotate it blind with a second model (GPT-5.4) under the same protocol. Dataset edges are excluded to match the model-to-model scope of the main lineage analysis. We score agreement at two levels: (i) \emph{exact-match agreement} at the repository level --- a repository matches only if the complete model-to-model dependency set is identical, with Cohen's $\kappa$ computed over complete dependency-set labels; and (ii) \emph{edge-level} precision, recall and $F_1$ on tuples $(\text{source}, \text{parent}, \text{type})$.

Overall exact-match agreement is 0.883 (53/60), with Cohen's $\kappa = 0.865$. Stratum-wise agreement is 0.821 for S1 Standard (23/28), 0.875 for S2 Rare (14/16) and 1.000 for S3 True negative (16/16); excluding true negatives, agreement is 0.841 (37/44). At the type-sensitive edge level, the re-annotation yields 39 true positives, 5 false positives and 7 false negatives, corresponding to precision 0.886, recall 0.848 and $F_1$ 0.867.

The seven disagreement cases were manually categorised against canonical Hugging Face tag conventions (e.g., \texttt{base\_model:finetune:<repo>} as evidence of a fine-tuning relationship) under the metadata-first evidence hierarchy. The disagreements are concentrated in boundary cases rather than reflecting different parent structures (Table~\ref{tab:s1_disagreement_categories}): three involve \texttt{base\_model:finetune} metadata being annotated as the coarser \texttt{base\_model}; one is a metadata--README conflict between quantization and fine-tuning; one is repository-identifier normalisation; and two are substantive annotation differences (missed or questionable model-parent evidence). This pattern supports use of the judge-derived annotations and motivates the separate reporting of topology, strict and type-compatible evaluation modes.

\begin{table}[h]
\footnotesize
\centering
\caption{Categorisation of the seven model-to-model disagreement cases in the independent re-annotation.}
\label{tab:s1_disagreement_categories}
\begin{tabular}{lcl}
\hline
Category & Count & Description \\
\hline
Coarse type assignment & 3 & \texttt{base\_model:finetune:<repo>} annotated as \texttt{base\_model} \\
Metadata--README conflict & 1 & Quantization metadata vs.\ fine-tuning prose \\
Repository-ID normalisation & 1 & Minor parent identifier string difference \\
Substantive annotation difference & 2 & Missed or questionable model-parent evidence \\
\hline
Total & 7 & -- \\
\hline
\end{tabular}
\end{table}

\section{Robustness of the merge-conflict association}
\label{supp:merge_robustness}

This section reports the identification strategy and robustness analyses underlying Section~2.3 of the main text. All analyses are restricted to cases whose direct dependencies include at least one merge edge and in which at least one scored parent carries ethical-use restriction evidence (score $\geq 1.0$). \emph{Conflict} cases have scored parents of mixed restrictive/non-restrictive intent; \emph{no-conflict} cases have all-restrictive scored parents. Parents with missing scores are excluded from treatment assignment but retained as the parent missing-rate covariate. The outcome is binary permissive relicensing (recognised permissive licence name, e.g.\ MIT, Apache-2.0, BSD-3-Clause); children with missing licence information are conservatively coded as non-relicensed.

\subsection{Raw and propensity-adjusted estimates}
\label{supp:merge_raw}

In the full sample ($n = 3{,}304$; $n_{\text{conflict}} = 2{,}289$, $n_{\text{no-conflict}} = 1{,}015$), permissive relicensing is more frequent under mixed-parent conflict than under all-restrictive parents ($p_{\text{conflict}} = 0.0660$ vs.\ $p_{\text{no-conflict}} = 0.0453$; $\Delta p = +0.0206$, 95\% bootstrap CI $[+0.0039, +0.0367]$, two-sided $z$-test $p = 2.08 \times 10^{-2}$).

\begin{figure}[h]
\centering
\includegraphics[width=0.7\textwidth]{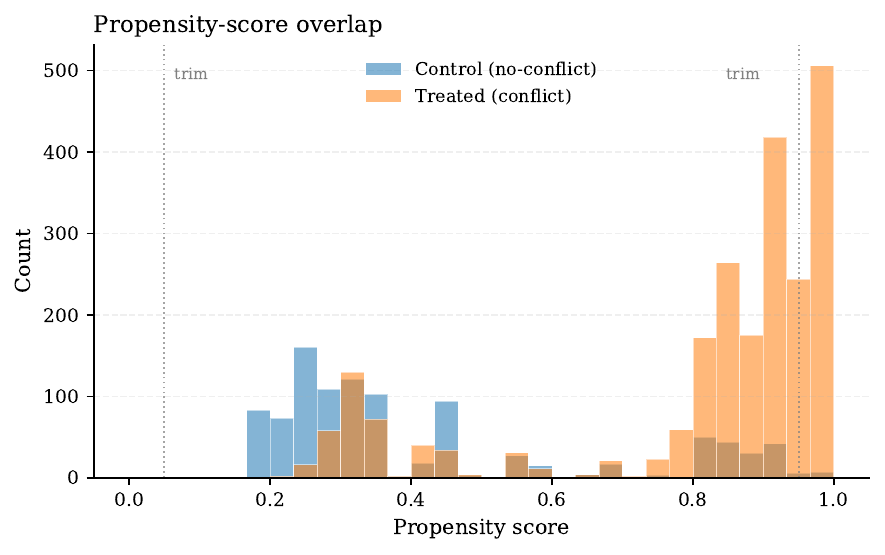}
\caption{\textbf{Propensity-score overlap between conflict and no-conflict merge groups.} Dotted lines indicate the trimming boundaries (0.05 and 0.95) used for IPW and AIPW.}
\label{fig:ps_overlap}
\end{figure}

Propensity scores are estimated by scaled logistic regression on nine pre-treatment features: number of parents, parent missing rate, log-transformed model age, log-transformed lineage size, their squared terms, and one parent-missing-rate $\times$ number-of-parents interaction. Parent restriction scores ($p_{\min}$, $p_{\max}$, $p_{\text{mean}}$) are excluded because they define treatment; downloads and likes are excluded as potential post-treatment variables. The propensity distributions for the two groups overlap in $[0.2, 0.5]$ but exhibit partial separation (Figure~\ref{fig:ps_overlap}): control units concentrate in the low-propensity region, treated units in the high-propensity region. This separation reflects genuine covariate differences (conflict merges have more parents and higher parent missing rates), motivating caliper-constrained matching and common-support trimming at $[0.05, 0.95]$.

\subsection{Matching, IPW and AIPW}
\label{supp:merge_psm}

Under 1:3 nearest-neighbour matching with caliper 0.20 and replacement, all 2,289 treated units match, drawing from 520 unique controls (out of 1,015 available); the matched ATT is $\Delta p = +0.0232$ (95\% clustered bootstrap CI $[+0.0119, +0.0348]$; 2,000 resamples clustered by treated unit). The PSM estimate is moderately sensitive to matching order under replacement, but the IPW and AIPW estimates on the trimmed sample are stable across reruns and directionally consistent.

\begin{figure}[h]
\centering
\includegraphics[width=0.65\textwidth]{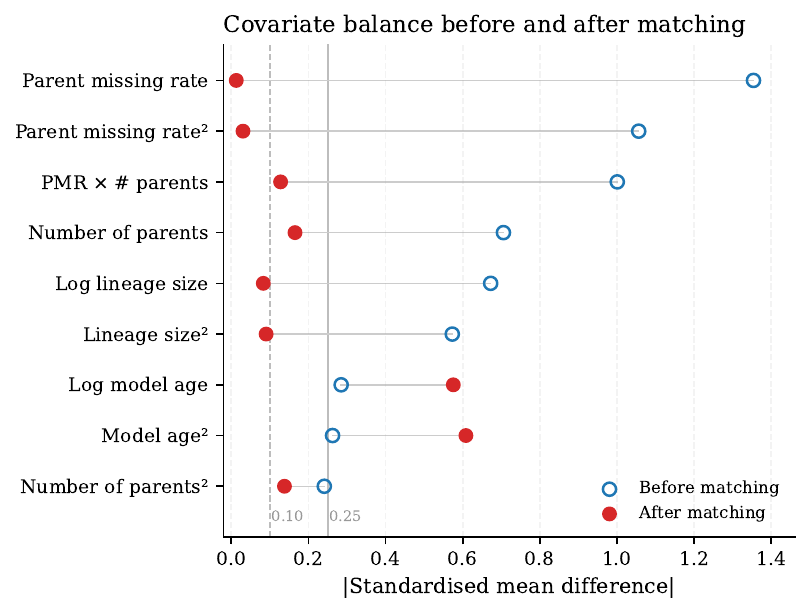}
\caption{\textbf{Covariate balance before and after propensity-score matching.} Open blue circles show $|\text{SMD}|$ before matching; filled red circles show $|\text{SMD}|$ after matching. Vertical reference lines mark the conventional 0.10 (dashed) and 0.25 (solid) thresholds. Matching improves balance on most covariates but worsens balance on model age, reflecting constraints imposed by the limited control pool under replacement matching.}
\label{fig:love_plot}
\end{figure}

Standardised mean differences before and after matching (Figure~\ref{fig:love_plot}) show that matching substantially improves balance on parent missing rate ($+1.35 \to +0.01$), number of parents ($+0.71 \to +0.17$) and log lineage size ($+0.67 \to +0.08$), all within the conventional 0.25 threshold. Residual imbalance persists on log model age ($+0.28 \to -0.58$) and its squared term ($+0.26 \to -0.61$): matching not only fails to improve balance on age but reverses its direction, because the limited control pool forces the algorithm to repeatedly draw from older controls. Unobserved confounders correlated with repository age --- evolving community norms, platform policy changes, temporal shifts in licence awareness --- could therefore partially account for the association.

On the trimmed sample, inverse-probability weighting (propensity refit after trimming) yields $\text{ATE} = +0.0287$ (bootstrap mean $+0.0282$, 95\% CI $[+0.0038, +0.0508]$; 800 resamples). Augmented inverse-probability weighting (AIPW)~\cite{robins1994estimation}, which combines the propensity model with separate scaled-logistic outcome regressions for treated and control on the trimmed sample, yields $\text{ATE} = +0.0294$ (bootstrap mean $+0.0288$, 95\% CI $[+0.0085, +0.0484]$; 800 resamples refitting both propensity and outcome on each resample).

\subsection{Summary and caveats}
\label{supp:merge_summary}

All four estimators agree in sign and order of magnitude (Table~\ref{tab:causal_summary}): the three adjusted estimates span $+0.023$ to $+0.029$ (AIPW at $+0.029$), while the raw difference is $+0.021$. This consistency across identification strategies strengthens the interpretation that the association is not solely driven by observable confounders.

\begin{table}[h]
\centering
\small
\caption{\textbf{Effect estimates across identification strategies.} PSM reports ATT; IPW and AIPW report ATE on the common-support-trimmed sample ($[0.05, 0.95]$).}
\label{tab:causal_summary}
\begin{tabular}{llcl}
\toprule
Estimator & Estimate & 95\% CI & Notes \\
\midrule
Raw $\Delta p$  & $+0.0206$ & $[+0.0039,\;+0.0367]$ & $p = 2.08 \times 10^{-2}$ \\
PSM (ATT)       & $+0.0232$ & $[+0.0119,\;+0.0348]$ & 1:3, caliper 0.20, replacement \\
IPW (ATE)       & $+0.0287$ & $[+0.0038,\;+0.0508]$ & Trimmed $[0.05, 0.95]$; propensity refit \\
AIPW (ATE)      & $+0.0294$ & $[+0.0085,\;+0.0484]$ & Doubly robust; refit bootstrap \\
\bottomrule
\end{tabular}
\end{table}

Three caveats apply. First, absolute relicensing probabilities are low in both groups (below 7\%), so the practical magnitude of the effect is modest. Second, residual imbalance on model age after matching (SMD $> 0.25$) means that unobserved confounders correlated with repository age cannot be fully excluded; stronger identification would require richer covariate sets (author-level controls, temporal fixed effects) or quasi-experimental variation in merge composition. Third, the partial separation in propensity distributions means IPW and AIPW rely on a relatively thin overlap region; while trimming at $[0.05, 0.95]$ mitigates extreme-weight bias, external validity beyond the overlap region warrants caution.
\section{Sensitivity of the governance horizon}
\label{supp:horizon_robustness}

This section examines robustness of the governance horizon $H^*$ (Methods Section~4.6) along three dimensions: the operational threshold $\alpha$, the merge-evidence threshold $\tau$, and the audit-state policy parameters.

\subsection{Operational threshold $\alpha$}
\label{supp:alpha_sweep}

\begin{figure}[h]
\centering
\includegraphics[width=0.6\textwidth]{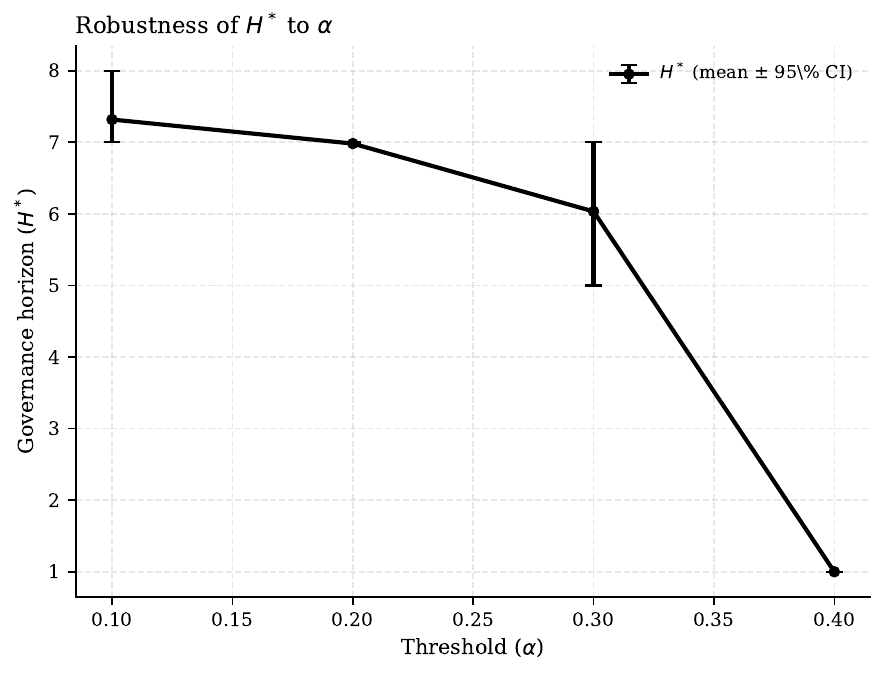}
\caption{\textbf{Robustness of the governance horizon to the operational threshold $\alpha$.} Error bars show 95\% bootstrap confidence intervals (500 resamples).}
\label{fig:alpha_sweep}
\end{figure}

$H^*(\alpha)$ is stable across reasonable thresholds (Figure~\ref{fig:alpha_sweep}, Table~\ref{tab:alpha_sweep}). At $\alpha = 0.10$ (requiring $\geq 90\%$ of downstream models to be non-auditable before declaring governance failure), $H^* = 7$~hops (bootstrap mean $7.32$, 95\% CI $[7, 8]$). At the primary $\alpha = 0.20$, $H^* = 7$~hops (mean $6.98$, CI $[7, 7]$). At $\alpha = 0.30$, $H^*$ shortens modestly to 6~hops (mean $6.03$, CI $[5, 7]$). At $\alpha = 0.40$, $H^* = 1$~hop, because $D(1) < 0.40$ already: more than 40\% of first-generation descendants lack sufficient public evidence for a governance determination. The last result underscores the severity of auditability collapse --- even tolerating 60\% undecidable cases, the horizon extends only one generation beyond the source.

\begin{table}[h]
\centering
\small
\caption{\textbf{Governance horizon under varying operational thresholds.} 500 bootstrap resamples.}
\label{tab:alpha_sweep}
\begin{tabular}{cccc}
\toprule
$\alpha$ & $H^*$ (point) & Bootstrap mean & 95\% CI \\
\midrule
0.10 & 7 & 7.32 & $[7,\;8]$ \\
0.20 & 7 & 6.98 & $[7,\;7]$ \\
0.30 & 6 & 6.03 & $[5,\;7]$ \\
0.40 & 1 & 1.00 & $[1,\;1]$ \\
\bottomrule
\end{tabular}
\end{table}

\subsection{Merge-evidence threshold and policy parameters}
\label{supp:tau_sweep}

The audit-state machine (Methods Section~4.5) has three configurable parameters: the merge-evidence threshold $\tau \in \{1, 2, 3, 4\}$ controlling when a mixed-parent merge is resolved as Decidable rather than UA; the reconciliation policy (strict, lenient, none) for resolved merge conflicts; and the upstream-missingness propagation policy (ambiguous, missing) for downstream nodes inheriting from a UM ancestor. We evaluate $H^*$ and $D(6)$ across the full $4 \times 3 \times 2 = 24$ combinations.

The results collapse to exactly two outcomes, determined entirely by the propagation policy (Table~\ref{tab:tau_sweep}): under the \emph{ambiguous} policy (UM propagates as UA), $H^* = 7$, $D(6) = 0.261$; under the \emph{missing} policy (UM propagates as UM), $H^* = 6$, $D(6) = 0.140$. $\tau$ and the reconciliation policy are inconsequential at ecosystem scale because the vast majority of governance-critical merges either have or lack sufficient evidence regardless of $\tau$. The propagation policy --- which determines whether the cascade of missing ancestors counts as ambiguity or missingness --- shifts $D(6)$ by $\sim$12 percentage points and $H^*$ by one hop. The main text uses the more conservative ambiguous policy ($H^* = 7$); under the less conservative missing policy, $H^* = 6$ remains shallow, confirming that the central finding does not hinge on this choice.

\begin{table}[h]
\centering
\small
\caption{\textbf{$H^*$ depends only on the upstream-missingness propagation policy.} All 12 combinations of $\tau \in \{1, 2, 3, 4\}$ and reconciliation policy $\in$ \{strict, lenient, none\} produce identical outcomes within each propagation policy. $n = 133{,}812$ reachable descendants; the full 24-row output is in the replication data (\texttt{rq4\_governance\_nmi\_tau\_robustness\_summary.csv}).}
\label{tab:tau_sweep}
\begin{tabular}{lcc}
\toprule
Upstream missing policy & $H^*$ & $D(6)$ \\
\midrule
Ambiguous (UM $\to$ UA downstream) & 7 & 0.261 \\
Missing (UM $\to$ UM downstream) & 6 & 0.140 \\
\bottomrule
\end{tabular}
\end{table}

\section{Platform intervention simulation}
\label{supp:intervention}

This section gives methodological details and full results for the platform intervention simulation reported in Section~3.1 of the main text. Starting from the baseline audit-state classification, we progressively resolve a fraction of Undecidable--Missing (U) nodes under one of three designs:

\begin{itemize}
\item \textbf{Inherit strictest}: a targeted U node inherits the strictest available upstream intent (R if any P/R ancestor is R, otherwise P); if no inheritable P/R ancestor exists, the node remains U.
\item \textbf{Auto-propagate}: inheritance applies only when all P/R ancestors agree; mixed-parent cases become UA, and orphan U nodes (no P/R ancestor) remain U.
\item \textbf{Mandatory licence declaration}: every targeted U node must receive a definite intent --- inheritance applies by default, and orphan U nodes fall back to a platform default of R. Only this design guarantees resolution of every targeted node.
\end{itemize}

The enforcement rate (the proportion of U nodes targeted) varies over 0\%, 25\%, 50\%, 75\% and 100\%. Stochasticity arises only from per-node Bernoulli draws over which U nodes are targeted at each rate; the lineage graph and SCC structure are fixed. We therefore report Monte Carlo simulation intervals over 500 stochastic enforcement realizations (Table~\ref{tab:intervention_results}). The evaluation window is 30 hops, with $H^* > 30$ right-censored.

\begin{table}[h]
\centering
\small
\caption{\textbf{Governance horizon under platform intervention scenarios.} 95\% Monte Carlo intervals over 500 enforcement realizations.}
\label{tab:intervention_results}
\begin{tabular}{lcccc}
\toprule
Intervention & Enforcement & Point $H^*$ & MC mean & MC interval \\
\midrule
None & 0\% & 7 & 7.000 & $[7,\;7]$ \\
Inherit strictest & 25\% & 7 & 7.000 & $[7,\;7]$ \\
Inherit strictest & 50\% & 7 & 7.024 & $[7,\;7]$ \\
Inherit strictest & 75\% & 7 & 8.526 & $[7,\;>30]$ \\
Inherit strictest & 100\% & $>30$ & $>30$ & $[>30,\;>30]$ \\
Auto-propagate & 25\% & 7 & 7.000 & $[7,\;7]$ \\
Auto-propagate & 50\% & 7 & 7.024 & $[7,\;7]$ \\
Auto-propagate & 75\% & 7 & 8.390 & $[7,\;>30]$ \\
Auto-propagate & 100\% & $>30$ & $>30$ & $[>30,\;>30]$ \\
Mandatory licence declaration & 25\% & 7 & 7.316 & $[7,\;9]$ \\
Mandatory licence declaration & 50\% & 7 & 8.516 & $[7,\;13]$ \\
Mandatory licence declaration & 75\% & 9 & 19.768 & $[8,\;>30]$ \\
Mandatory licence declaration & 100\% & $>30$ & $>30$ & $[>30,\;>30]$ \\
\bottomrule
\end{tabular}
\end{table}

The results separate into two regimes. Inheritance-only designs (inherit-strictest, auto-propagate) leave $H^*$ unchanged at 7 hops through 50\% enforcement, enter a transitional regime at 75\% with Monte Carlo intervals spanning $[7, >30]$ depending on which subset is resolved, and only push $H^* > 30$ at 100\% enforcement. Mandatory licence declaration, which forces a definite intent on orphan U nodes, moves $H^*$ at intermediate rates --- mean $H^*$ reaches $7.32$ at 25\%, $8.52$ at 50\% and $19.77$ at 75\%. The binding structural constraint is therefore not enforcement intensity alone but the presence of orphan-U lineage components; policies that explicitly resolve such components can extend the governance horizon at moderate enforcement, though deep auditability still requires near-complete coverage.

\section{PyPI comparator ecosystem}
\label{supp:pypi}

\subsection{Data, sampling and LRI mapping}
\label{supp:pypi_data}

We use a PyPI release-metadata snapshot of package names, versions, declared dependencies and licences. For each release we extract the dependency tree from structured metadata (\texttt{dep\_tree\_created}) and construct a directed graph from upstream dependencies to downstream dependants. Because frequently updated packages would otherwise dominate the hop-distance distribution, we apply time-stratified sampling: for each (package, calendar year) pair we retain at most one release, chosen by deterministic hashing of \texttt{name|year|version} to ensure reproducibility.

Each package's licence is mapped to a licence restrictiveness index (LRI) on $[0, 1]$ using a reference table of 63 licence terms graded by copyleft strength: strong copyleft (GPL-family, AGPL-3.0; LRI $=1.0$), moderate copyleft (LGPL family; $0.75$), weak / file-level copyleft (MPL-2.0, CDDL, EPL-2.0; $0.50$), and permissive (MIT, Apache-2.0, BSD, ISC, Unlicense; $0.0$). Licence strings are matched by exact lookup on the normalised name, then by substring matching for non-standard declarations; packages with no resolvable licence are excluded. The complete 63-entry mapping is in Table~\ref{tab:lri_mapping} and as \texttt{licenses\_terms\_63.csv} in the replication data.

\begin{table}[h]
\centering
\scriptsize
\caption{\textbf{Licence restrictiveness index mapping for the PyPI comparator.}}
\label{tab:lri_mapping}
\begin{tabular}{lp{0.76\textwidth}}
\toprule
LRI & Licences \\
\midrule
1.00 & AGPL-3.0; AGPL-3.0+; CC-BY-SA-4.0; CECILL-2.1; EPL-1.0; EUPL-1.1; EUPL-1.2; GPL-2.0; GPL-2.0+; GPL-3.0; GPL-3.0+; LiLiQ-Rplus-1.1; MS-PL; MulanPubL-2.0; NPOSL-3.0; OSL-3.0; RPL-1.5; SimPL-2.0 \\
0.75 & LGPL-2.1; LGPL-2.1+; LGPL-3.0; LGPL-3.0+; OGTSL \\
0.50 & CDDL-1.0; CPAL-1.0; EPL-2.0; LiLiQ-R-1.1; MPL-2.0; MS-RL; NOSL \\
0.00 & 0BSD; AAL; AFL-3.0; Apache-2.0; Artistic-2.0; BSD-1-Clause; BSD-2-Clause; BSD-2-Clause-Patent; BSD-3-Clause; BSD-3-Clause-Clear; BSD-4-Clause; BSL-1.0; CC-BY-4.0; CC0-1.0; ClArtistic; ECL-2.0; EFL-2.0; Fair; FSFAP; Imlib2; ISC; LiLiQ-P-1.1; MirOS; MIT; MIT-0; MulanPSL-2.0; NCSA; NTP; Ruby; Unlicense; UPL-1.0; WTFPL; Zlib \\
\bottomrule
\end{tabular}
\end{table}

\subsection{Per-hop statistics and limitations}
\label{supp:pypi_perhop}

Source packages are dependency roots (in-degree zero) with LRI $= 1.0$ (GPL-family); downstream distance is shortest-path distance from the nearest GPL source, with a 6-hop cutoff. Table~\ref{tab:pypi_perhop} reports the mean downstream LRI and sample size at each hop. Sample sizes decrease from 80,423 at hop~0 to 80 at hop~6; the apparent upturn at hop~6 is based on only 80 observations and is not statistically distinguishable from the hop~3--5 plateau. An exponential decay model fits poorly ($R^2 = 0.13$, $T_{1/2} = 6.74$ hops), in sharp contrast to the Hugging Face ecosystem ($R^2 = 0.98$): PyPI restrictiveness stabilises after an initial drop rather than collapsing monotonically toward zero. This contrast underpins the main-text claim that the governance horizon is topology-specific rather than generic to open ecosystems.

\begin{table}[h]
\centering
\small
\caption{\textbf{Per-hop statistics for the PyPI comparator ecosystem.} Mean LRI by dependency distance from GPL source packages; 95\% CI based on the normal approximation. Statistics use year-stratified sampling (at most one release per package per calendar year).}
\label{tab:pypi_perhop}
\begin{tabular}{cccc}
\toprule
Hop & $n$ & Mean LRI & 95\% CI \\
\midrule
0 & 80,423 & 1.000 & --- \\
1 & 67,890 & 0.530 & $[0.526,\;0.534]$ \\
2 & 34,412 & 0.420 & $[0.414,\;0.426]$ \\
3 & 12,027 & 0.317 & $[0.309,\;0.325]$ \\
4 & 2,594  & 0.302 & $[0.285,\;0.320]$ \\
5 & 524    & 0.421 & $[0.380,\;0.462]$ \\
6 & 80     & 0.784 & $[0.696,\;0.872]$ \\
\bottomrule
\end{tabular}
\end{table}

The comparator is not designed to equate models with software packages legally or technically; the two ecosystems differ across many dimensions (copyleft enforcement vs.\ voluntary disclosure; machine-readable manifests vs.\ textual model cards; code linking vs.\ weight-level transformation). What it tests is whether the depth-bounded auditability collapse observed in open-weight lineages is generic to open ecosystems or specific to weight-level reuse, by holding constant only the abstract structure (rooted dependency graph + licence-graded nodes) while varying the propagation mechanism. Two limitations apply: PyPI measures restrictiveness (continuous) rather than auditability (binary), because PyPI packages rarely have missing licences, so the comparison tests whether governance \emph{signals} degrade with depth rather than whether \emph{auditability} collapses; and the hop-6 sample size ($n = 80$) is small, so the apparent upturn at that depth should not be over-interpreted.

\section{Licence gold-set construction and classifier validation}
\label{supp:license_goldset}

This section reports the gold-standard annotation procedure used to validate the ethical-use restriction classifier (Methods Section~4.4).

\subsection{Sampling, codebook and adjudication}
\label{supp:license_sampling}

The gold set contains 200 licence texts sampled from the Hugging Face licence corpus by a hybrid frequency strategy: standard licences with recognised identifiers are aggregated by name, custom or \texttt{other} licences by exact text content, and the resulting groups are ranked by ecosystem-wide frequency. The 200 highest-frequency groups form the evaluation sample, with one representative full licence text retained per group.

Each licence was independently coded by GPT-5 and by a domain-informed human annotator under the same structured codebook. The primary binary variable is whether the licence contains ethical or behavioural use restrictions (e.g., prohibitions on violence, hate, disinformation, surveillance, malware, high-risk decision-making, or other harmful-use domains); purely permissive, copyleft-only, and commercial/research-restricted licences are coded as non-ethical. Auxiliary variables (broad category, anti-distillation, commercial restrictions, copyleft status, ethical-use domains mentioned) are recorded for error analysis but not used for classifier validation. Disagreements were resolved by discussion-based adjudication under a protocol that counts mandatory acceptable-use policies incorporated by reference as licence-layer ethical-use restriction evidence, even when the policy is linked rather than reproduced.

The rule-based classifier (Table~\ref{tab:license_classifier_rules}) normalises licence text, identifies restriction-context windows around headings such as \emph{use restrictions}, \emph{prohibited uses}, \emph{safety}, \emph{compliance} and \emph{acceptable use}, and scores behavioural restriction evidence only when harm-domain terms occur within a prohibition or restriction context. This context gating suppresses false positives from generic legal-compliance language.

\begin{table}[h]
\centering
\scriptsize
\caption{\textbf{Pattern inventory and scoring scheme for the ethical-use restriction classifier.} A licence is classified as containing ethical-use restriction evidence when the total score is at least 1.0.}
\label{tab:license_classifier_rules}
\begin{tabular}{p{0.19\textwidth}p{0.47\textwidth}p{0.22\textwidth}}
\toprule
Rule class & Pattern examples & Score contribution \\
\midrule
Restriction frame & must not; shall not; may not; prohibited; not permitted; restricted; forbidden; banned; you agree not to; not allowed & Defines sentence-level restriction context \\
Policy reference & acceptable use policy; AUP; prohibited use policy; use policy; responsible use; safety policy; usage guidelines; content policy & $+1.0$ \\
Section hint & restrictions; rules of use; use restrictions; prohibited uses; safety; compliance; acceptable use & Opens an 800-character restriction-context window \\
Hard harm domains & military; weapon; nuclear; terrorism; surveillance; biometric; facial recognition; law enforcement; CSAM; hate speech; discrimination; malware; phishing; deepfake; election; high-risk use; social scoring; credit; employment; housing; medical; legal advice; financial advice & $1.5 + 0.5 \times$ additional hits, capped at $+2.5$, when in restriction context \\
Soft harm domains & harmful; misinformation; disinformation; deceptive; misleading; harassment; defamation; fraud; scam; spam; privacy; personal data; PII; vulnerability; exploitation & $+0.4$ per hit, capped at $+1.2$, when in restriction context \\
Brand and policy context & Gemma; Llama; OpenRAIL/RAIL; CreativeML; Stable Diffusion; BigScience; Mistral; Falcon; Qwen; Hunyuan, combined with terms, agreement, licence, policy or URL context & $+0.8$ unless an exclusion phrase applies \\
Exclusions & not affiliated with; not governed by; not a product of; Google claims no rights & Suppresses brand-only evidence \\
\bottomrule
\end{tabular}
\end{table}

\subsection{Agreement and classifier validation}
\label{supp:license_validation}

Before adjudication, the independent GPT-5 and human labels agreed on 191 of 200 licences (agreement $= 0.955$; Cohen's $\kappa = 0.908$). The 9 disagreements were adjudicated as 7 retaining the LLM label and 2 retaining the human label; the resulting gold set contains 89 ethical-use-restriction licences and 111 non-ethical licences.

Against the adjudicated gold set, the rule-based classifier correctly identifies 81 of 89 ethical-use restriction licences (TP $= 81$, FN $= 8$) and 108 of 111 non-ethical licences (TN $= 108$, FP $= 3$), giving precision $= 0.964$, recall $= 0.910$, $F_1 = 0.936$ and accuracy $= 0.945$ for the ethical-use restriction class. Per-class metrics for the non-ethical class are precision $= 0.931$, recall $= 0.973$, $F_1 = 0.952$ (support 111); for the ethical class, the metrics above (support 89).

\end{document}